# Explicit Learning: an Effort towards Human Scheduling Algorithms

**Jingpeng Li, Uwe Aickelin**
School of Computer Science, University of Nottingham, NG8 1BB UK,
uxa@cs.nott.ac.uk

## Abstract

Scheduling problems are generally NP-hard combinatorial problems, and a lot of research has been done to solve these problems heuristically. However, most of the previous approaches are problem-specific and research into the development of a general scheduling algorithm is still in its infancy.

Mimicking the natural evolutionary process of the survival of the fittest, Genetic Algorithms (GAs) have attracted much attention in solving difficult scheduling problems in recent years. Some obstacles exist when using GAs: there is no canonical mechanism to deal with constraints, which are commonly met in most real-world scheduling problems, and small changes to a solution are difficult. To overcome both difficulties, indirect approaches have been presented (in [1] and [2]) for nurse scheduling and driver scheduling, where GAs are used by mapping the solution space, and separate decoding routines then build solutions to the original problem.

In our previous indirect GAs, learning is implicit and is restricted to the efficient adjustment of weights for a set of rules that are used to construct schedules. The major limitation of those approaches is that they learn in a non-human way: like most existing construction algorithms, once the best weight combination is found, the rules used in the construction process are fixed at each iteration. However, normally a long sequence of moves is needed to construct a schedule and using fixed rules at each move is thus unreasonable and not coherent with human learning processes.

When a human scheduler is working, he normally builds a schedule step by step following a set of rules. After much practice, the scheduler gradually masters the knowledge of which solution parts go well with others. He can identify good parts and is aware of the solution quality even if the scheduling process is not completed yet, thus having the ability to finish a schedule by using flexible, rather than fixed, rules. In this research we intend to design more human-like scheduling algorithms, by using ideas derived from Bayesian Optimization Algorithms (BOA) and Learning Classifier Systems (LCS) to implement explicit learning from past solutions.

BOA can be applied to learn to identify good partial solutions and to complete them by building a Bayesian network of the joint distribution of solutions [3]. A Bayesian network is a directed acyclic graph with each node corresponding to one variable, and each variable corresponding to individual rule by which a schedule will be constructed step by step. The conditional probabilities are computed according to an initial set of promising solutions. Subsequently, each new instance for each node is generated by using the corresponding conditional probabilities, until values for all nodes have been generated. Another set of rule strings will be generated in this way, some of which will replace previous strings based on fitness selection. If stopping conditions are not met, the Bayesian network is updated again using the current set of good rule strings. The algorithm thereby tries to explicitly identify and mix promising building blocks. It should be noted that for most scheduling problems the structure of the network model is known and all the variables are fully observed. In this case, the goal of learning is to find the rule values that maximize the likelihood of the training data. Thus learning can amount to 'counting' in the case of multinomial distributions.

In the LCS approach, each rule has its strength showing its current usefulness in the system, and this strength is constantly assessed [4]. To implement sophisticated learning based on previous solutions, an improved LCS-based algorithm is designed, which consists of the following three steps. The initialization step is to assign each rule at each stage a constant initial strength. Then rules are selected by using the Roulette Wheel strategy. The next step is to reinforce the strengths of the rules used in the previous solution, keeping the strength of unused rules unchanged. The selection step is to select fitter rules for the next generation. It is envisaged that the LCS part of the algorithm will be used as a hill climber to the BOA algorithm.

This is exciting and ambitious research, which might provide the stepping-stone for a new class of scheduling algorithms. Data sets from nurse scheduling and mall problems will be used as test-beds. It is envisaged that once the concept has been proven successful, it will be implemented into general scheduling algorithms. It is also hoped that this research will give some preliminary answers about how to include human-like learning into scheduling algorithms and may therefore be of interest to researchers and practitioners in areas of scheduling and evolutionary computation.

## References

1. Aickelin, U. and Dowsland, K. (2003) "An Indirect Genetic Algorithm for a Nurse Scheduling Problem", Computer & Operational Research (in print).
2. Li, J. and Kwan, R.S.K. (2003), "A Fuzzy Genetic Algorithm for Driver Scheduling", European Journal of Operational Research 147(2): 334-344.
3. Pelikan, M., Goldberg, D. and Cantu-Paz, E. (1999) "BOA: The Bayesian Optimization Algorithm", IlliGAL Report No 99003, University of Illinois.
4. Wilson, S. (1994) "ZCS: A Zeroth-level Classifier System", Evolutionary Computation 2(1), pp 1-18.